%% file: paper.tex
\documentclass[letterpaper, 10pt, conference]{ieeeconf}  

\IEEEoverridecommandlockouts                              
\usepackage{graphicx} 
\usepackage[tight,footnotesize]{subfigure}
\usepackage{epsfig} 
\usepackage{amsmath} 
\usepackage{amssymb}  
\usepackage{hyperref}
\usepackage{xcolor}
\usepackage{booktabs}
\usepackage{pgfplots}
\usepackage{csvsimple}
\pgfplotsset{compat=1.6}
\usepackage{caption}
\hypersetup{
	colorlinks,
 	linkcolor={black},
 	citecolor={black},
 	urlcolor={black}
}

\title{Reinforcement Learning for Agile Active Target Sensing with a UAV}

\author{Harsh Goel$^{1}$, Laura Jarin Lipschitz$^{1}$, Saurav Agarwal$^{1}$, Sandeep Manjanna$^{2}$, and Vijay Kumar$^{1}$
\thanks{We gratefully acknowledge the support from ARL Grant DCIST CRA
W911NF-17-2-0181, NSF Grant CCR-2112665, CNS-1446592, AND and EEC-1941529,
Qualcomm Research,NVIDIA and C-BRIC, a Semiconductor Research Corporation Joint University Microelectronics Program program cosponsored by
DARPA.}
\thanks{$^{1}$GRASP Laboratory, University of Pennsylvania, PA, USA (\{harshg99, laurajar, kumar\}@seas.upenn.edu)}
\thanks{$^{2}$Plaksha University, Mohali, India 
(sandeep.manjanna@plaksha.edu.in)}
}

\begin{document}
\maketitle

\begin{abstract}

Active target sensing is the task of discovering and classifying an unknown number of targets in an environment and is critical in search-and-rescue missions. This paper develops a deep reinforcement learning approach to plan informative trajectories that increase the likelihood for an uncrewed aerial vehicle (UAV) to discover missing targets. Our approach efficiently (1) explores the environment to discover new targets, (2) exploits its current belief of the target states and incorporates inaccurate sensor models for high-fidelity classification, and (3) generates dynamically feasible trajectories for an agile UAV by employing a motion primitive library. Extensive simulations on randomly generated environments show that our approach is more efficient in discovering and classifying targets than several other baselines. A unique characteristic of our approach, in contrast to heuristic informative path planning approaches, is that the it is robust to varying amounts of deviations of the prior belief from the true target distribution, thereby alleviating the challenge of designing heuristics specific to the application conditions.

\end{abstract}

\section{Introduction}
\label{sc:intro}
\input{intro}
%


\section{Problem Description}
\label{sc:problem}
\input{problem_desc}

\section{Related Work}
\label{sc:related_work}
\input{related_work}

\section{Target Mapping Approach}
\label{sc:mapping}
\input{mapping}


\section{Planning Approach}
\label{sc:approach}
\input{rl_approach}

\section{Results}
\label{sc:sim}
\input{results}

\section{Conclusion}
\input{conclusion}

\bibliographystyle{IEEEtran}
\bibliography{ipp}
\end{document}

%% file: intro.tex
Consider a search and rescue scenario where an uncrewed aerial vehicle (UAV) is deployed after an earthquake to gather information using its sensors to identify and locate trapped human beings. For search, a {\em prior} over where these human targets can be through correlated information such as infrastructural damage through satellite images. The following question then arises: How should we plan UAV trajectories to gather essential information for locating humans while respecting the dynamics and the endurance of the UAV? This problem belongs to the broad class of NP-hard {\em informative path planning} (IPP) problems~\cite{KrauseThesis}.

This paper addresses several challenges in planning a reliable and efficient search trajectory for a UAV. First, the trajectory needs to respect the dynamics and battery capacity of the UAV to ensure a feasible, smooth, and efficient traversal in the environment. Second, the prior need not be accurate, may have outdated data, and may be based on correlated phenomenon instead of the actual phenomenon of interest. This leads to the classical trade-off between {\em exploration} and {\em exploitation} where balancing exploration (which involves sensing regions of high uncertainty or missing data) and exploitation (to confirm existence of features of interest under budget constraints) and sensing uncertainty is a difficult problem.

\begin{figure}
    \centering
    \includegraphics[width=3.25in,height=1.5in]{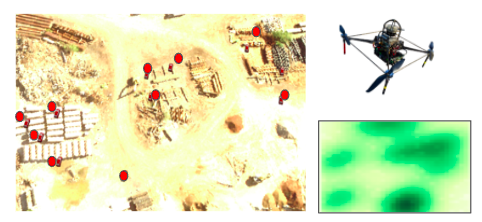}
    \caption{Quadrotor in a Search and Rescue setting, image on the left shows a disaster site with rubble and targets in red and the quad on the right has a prior belief over these targets}
    \label{quadrotor_sar}
\end{figure}
This paper develops a reinforcement learning (RL) framework for informative path planning with the following salient features:
\begin{enumerate}
	\item {\em Dynamic feasibility:} The framework generates policies with dynamically feasible motion primitives to achieve smooth and stable trajectories.
	\item {\em Online adaptive model:} The framework updates its current belief of the environment during the execution of the robot path and uses this updated belief to generate further policies.
	\item {\em Online decision-making:} Robust decision making that trades-off between exploration and exploitation over belief priors with varying confidence.
\end{enumerate}


%% file: problem_desc.tex
The active target sensing problem is cast into an informative trajectory planning problem where the overall objective is to choose an optimal trajectory $\psi^*$ from a collection of admissible trajectories $\mathbf{\Psi}$ that maximizes the information theoretic utility $I$ on the target estimate. The IPP objective is defined as \cite{PopovicHNSSG17}
\begin{equation}
        \psi^* = \underset{\psi \in \mathbf{\Psi}}{\mathrm{argmax}} \;I [Measure(P)] 
\end{equation}
\begin{equation*}
         s.t\; Energy(P) \leq B 
\end{equation*}

Where B denotes the energy budget available for the flight of the drone, MEASURE (.) denotes the change in estimates from discrete measurements taken along the path $\psi$ and ENERGY (·) provides the energy cost of the path taken.

%% file: related_work.tex
Informative path planning (IPP) is a broad class of problems with the primary goal of collecting information from the environment. These methods have been broadly applied for many active sensing problems such as target search \cite{PopovicHNSSG17,MeeraPMS19,viseras2019deepig}, environmental monitoring \cite{moon2022tigris,RuckinJP22} and mapping \cite{PopovicVCNS20,schmid2020efficient}. 

IPP algorithms for specific applications can be broadly classified into 4 categories: sampling, optimization, combinatorial and learning based. Combinatorial methods \cite{SuhKcho16,binney2012branch} are inefficient for online planning as these methods query an exponentialy large search space to obtain feasible trajectories.  Sampling \cite{PopovicHNSSG17,MeeraPMS19,PopovicVCNS20} and optimization \cite{moon2022tigris,schmid2020efficient,HollingerS14,vivaldini2019uav} based approaches improve over these methods by sampling a sequence of way-points or control actions to compute candidate paths and have shown to be probabilistically optimal. However, these methods can be computationally inefficient for online planning during agile flights especially due to the forward propagation of many simulated measurements across many sampled candidate paths. 
Given the advances in Reinforcement learning (RL) \cite{arulkumaran2017deep}, many RL based IPP solutions have been formulated for exploration \cite{viseras2019deepig}, environmental monitoring \cite{RuckinJP22}, search and rescue\cite{niroui2019deep} and localization of sources \cite{wiedemann2021robotic}. However, these methods have been shown to be restricted to small action spaces of next best measurement sites, hence trajectory planning is dynamics in-aware.
 
In contrast to these methods, our approach employs a set of dynamically feasible motion primitives that are computed offline. This enables online informative trajectory generation for agile UAVs. Inspired by the recent advances in RL, our algorithm leverages the A3C method to train a policy over a library of motion primitives that plans dynamically feasible informative paths.

%% file: mapping.tex
This section describes the approach to mapping targets. The map is modeled via an occupancy grid that is updated via measurements from a modeled camera sensor.

\subsection{Sensor Model}

The UAV is equipped with a downward looking camera that provides a binary measurement  $z_t \; \in \; [0,1]$ at time $t$ that classifies whether the observed cell has a target $z_t = 1$. The sensor has a fixed rectangular footprint on the environment and the likelihood of the sensor providing correct measurements $p(z_t|m_{xy})$ given whether a cell has a target or is free, varies with the distance of the camera.



\subsection{Environment Model}
\label{env_model}
A UAV maintains a 2D discrete occupancy grid over the search environment. Each cell indicates the probability of a target.  A measurement $z_t$ at a given time $t$ updates the observed cell $m_{xy}$ via the log updates \cite{ThrunOcc} as follows: 
\begin{equation}
 L (m_{xy}|z_{1:t}, x_{1:t}) = \sum_{t=1}^{N} log \frac{p(z_t|m_{xy})}{p({z_t|\Tilde{m}_{xy})}}  + L(m_{xy})   
\end{equation}

where $p(z_t|m_{xy})$ denotes the forward sensor model, $m_{xy}$ refers to an occupied cell and $\Tilde{m}_{xy}$ refers to a free cell,and $L(m_{xy})$ denotes the log odds of the prior belief at location $(x,y)$ in the grid. A target is said to be found if the occupancy is above 95\%.

Hence the UAV is tasked to minimise the map uncertainty while maximising certainty of targets maintained via the environment model and measured via the sensor model.

%% file: rl_approach.tex

The proposed planner with motion primitives is a different approach to informative path planning for target search in environments which would require agility and high speed. Our approach has two key features. First, this method exploits the strengths of motion primitives that respect the vehicle dynamics and can be extended spatially to plan in unbounded configuration spaces. Second, this approach learns a policy network network over motion primitives conditioned on scaled up egocentric views of feature maps. The policy can be executed during run time for fast re-planning based on the current state of the occupancy map.



\subsection{Trajectory Parameterization via Dispersion Minimizing Motion Primitive Graphs}

By incorporating dynamics information, UAV's can execute agile paths for a monitoring or target search task which is a primary limitation of prior work. As such we parameterize trajectories by generating a motion primitive graph over sampled states that minimize the asymetric dispersion costs \cite{laurajar1} evaluated via trajectory optimization assuming quadrotor dynamics\cite{liu2017search} .  
\begin{equation}
        d(V) = \sup_{\mathbf{x} \in X} \min_{\mathbf{v} \in V}[\max(J(\mathbf{x}, \mathbf{v}), J(\mathbf{v}, \mathbf{x}))]
\end{equation}

Hence, our approach defines these trajectory segments offline and can be spatially extended by leveraging the affine property of UAV dynamics for planning in unbounded configuration spaces \cite{laurajar2}. 

\subsection{IPP with primitives as a RL problem}
We cast IPP problem as an RL problem where the objective is to maximize the total information gain for target classification over a trajectory sampled from the spatially extended motion primitive graph. The POMDP for the IPP is defined as follows:

\subsubsection{States}
\label{states}
In addition to the target occupancy, agents maintain an obstacle map and a coverage map. The three maps and the quadrotor's global state for the full state of the POMDP as shown in Fig \ref{architecture}.

\subsubsection{Observations}
For informative search applications, the size of the environment may vary by application. 
Thus, we adopt an egocentric view of the information map for the agent. This egocentric view is scaled up albeit with different resolutions (see fig \ref{architecture}). 
In addition, the position of the UAV normalized to the size of the map, the one hot encoding of the prior action taken and the current node $V$ on the motion primitive graph $G$ are added as observations.

\subsubsection{Actions}
\label{actions}
Given a motion primitive graph $(G = {V,E})$, the action space of the UAV is composed of primitives $E(\mathbf{v})$ at the node $\mathbf{v}$ of the motion primitive graph. A stochastic policy is outputted over the set of valid motion primitives at the given state. These valid primitives are learned for collision avoidance through a valid loss specified in \ref{learning}.

\subsubsection{Rewards}
Rewards are designed for the agents to reduce in uncertainty of the target occupancy map, prioritize covering high target occupancy estimate regions and find targets as quickly as possible.

The uncertainty in the target occupancy map is characterized by it's total Shannon's entropy as follows:
\begin{equation}
\mathcal{H}(M_t) =  \sum_{xy \in \mathbb{R}^2} \mathcal{H}(m_{xy}\mid z_{1:t})
\end{equation}

The rewards due to the reduction in the uncertainty of the map is given by:
\begin{equation}
r_t = \mathcal{H}(M_{t+1}) -\mathcal{H}(M_{t})
\end{equation}
where $M_{t+1}$ is the state of the map at time $t+1$ after primitive action $a_{t}$ is executed.

The coverage reward is defined as the sum of the target probabilities covered by if primitive $a_t$ was executed. This reward is essential to find highly correlated targets as quickly as possible under limited budgets.
Agents are rewarded for finding targets once the occupancy probability exceeds 0.95. The total reward for finding all targets is 100 and the number of targets can vary with the training map.
In addition, a penalty is in proportion to the cost of the selected primitive. This cost is quantified by the total energy cost and time taken to execute the primitive \cite{laurajar1}.

\subsection{Learning}
\label{learning}



%
We leverage the A3C framework to train a policy over the motion primitive graph where environments are generated across many local worker threads and a global policy network is updated via gradients from local experience buffers.\cite{mnih2016asynchronous}. 

\subsubsection{Environment Generation}

During each episode, targets are generated from a ground truth distribution (GT) (referred to as world model) which is modeled as a GMM with sampled centers and diagonal co-variances. Map generation prevents over-fitting to a single scenario. An agent is only given a prior over its target occupancy map which is initialized via a combination of the following: 
\begin{enumerate}
    \item Shifting the centers of the original GT   .
    \item Sampling GMM noise and adding it to the GT.
    \item Adding discrete Gaussian noise to the GT.
\end{enumerate} 

\subsubsection{Network Architecture}

\begin{figure}
    \centering
    \includegraphics[width=3in,height=2in]{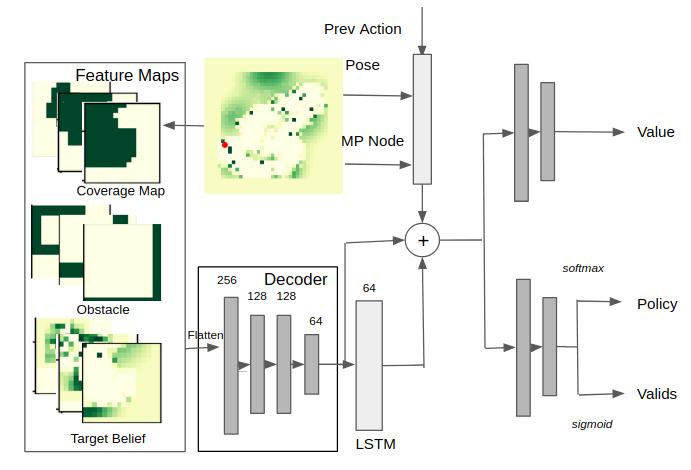}
    \caption{Model Architecture}
    \label{architecture}
\end{figure}

The actor-critic network outputs a policy $\boldsymbol{\pi}(s_t;\theta_c)$, value $V(s_t;\theta_c)$ and valids $\boldsymbol{\psi}(s_t;\theta_c)$ as shown in Fig \ref{architecture} from the observation inputs.
\subsubsection{Training Losses}
\label{loss}
The agent samples an action (primitive) using the policy distribution and an episode terminates once the agent exhausts its budget.

The critic is regressed against the discounted lambda returns $G_t^{\lambda} $\cite{sutton1998introduction}. Empirically $\lambda = 0.8$ gave us the best results. 
The critic loss is given by 
\begin{gather}
  L_{\theta_c} = \sum_{t=0}^{T} (G_t^{\lambda} - V(s_t;\theta_c))^2  
\end{gather}

To improve the policy $\pi(a_t|s_t;\theta_p)$, the advantage $A(s_t,a_t;\theta) = r(s_t,a_t) + \gamma V(s_{t+1};\theta_c) - V(s_t;\theta_c)$ is computed for the current action.
These advantage terms are discounted from a given time step to compute the generalised advantage estimate $A_t^{GAE}$ for the loss $L_{actor} = \sum _{t=1}^T log\;(\pi(a_t|s_t;\theta_p))\;A_t^{GAE}$.
In addition, we add a one step entropy loss $L_{ent}= \sum_{t=1}^T \mathcal H(\boldsymbol{\pi}(s_t);\theta_p))$ to encourage exploration and a supervised asymmetric binary cross entropy loss $L_{val}$ to ensure that the policy network outputs a valid distribution.
\begin{equation}
    L_{val} = \sum_{t=1}^{T} \beta_1 (1-\mathbf{y_t})log (1-\boldsymbol{\phi}(s_t;\theta_p)) +
\beta_2 \mathbf{y_t} log(\boldsymbol{\phi}(s_t;\theta_p))
\end{equation}
where $\mathbf{y_t}$ refers to the set of valid actions/primitives.
In summary the policy loss can be given by:
\begin{align}
  L(\theta_p) = \alpha_1 L_{ent} - \alpha_2\; L_{actor} + L_{valid}  
\end{align}

%% file: results.tex
\begin{figure*}[htp]
    \centering
	\subfigure[]{ \includegraphics[height=1.3in]{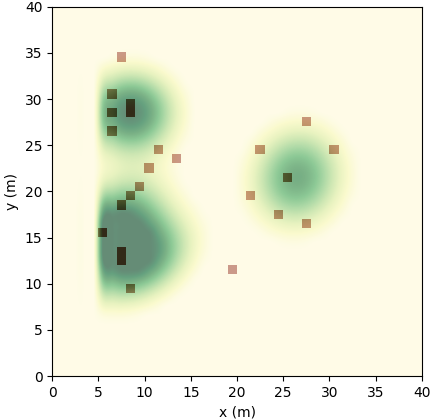}}
	\hspace{-1em}
	\subfigure[]{ \includegraphics[height=1.3in]{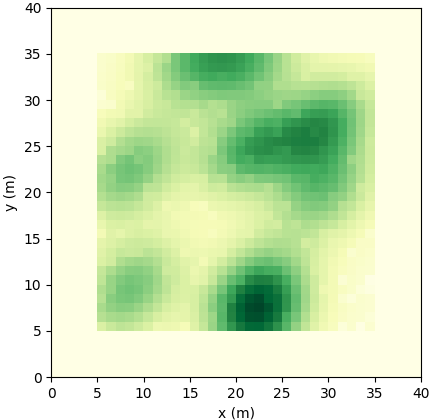}}
	\hspace{-1em}
	\subfigure[]{ \includegraphics[height=1.3in]{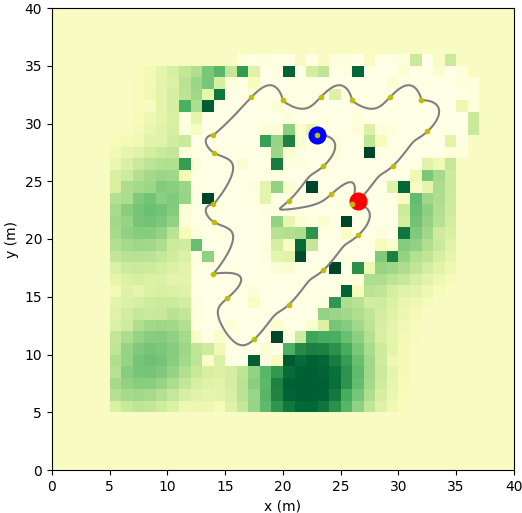}}
	\hspace{-1em}
	\subfigure[]{ \includegraphics[height=1.3in]{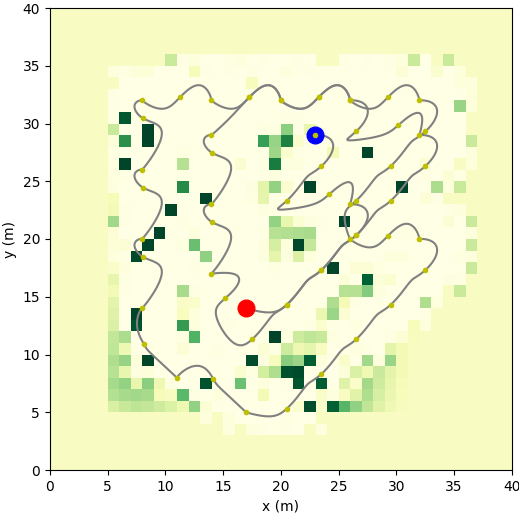}}
	\hspace{-1em}
	\subfigure[]{ \includegraphics[height=1.37in]{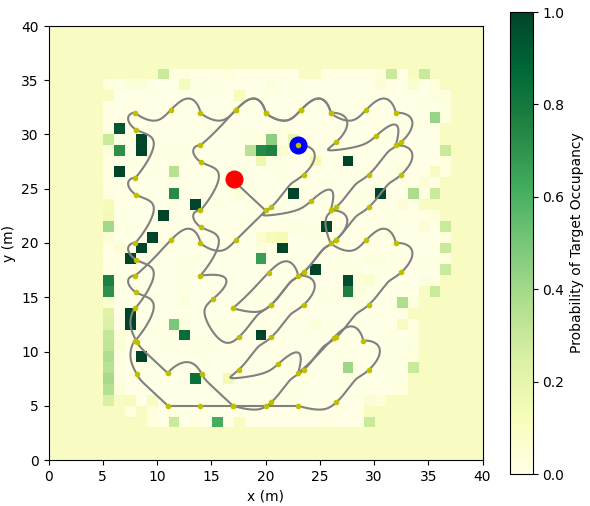}}
    \caption{Visualization of the traveled trajectory at varying remaining budgets. Agent's starting position is depicted via a green dot and the ending position is depicted via a red dot. From right to left, the plots show the (a) True target distribution with targets marked in red; (b) Prior belief of the map; the following plots show the travelled trajectory at c) 70\% budget; d) 30\% budget; (e) 0\% budget.} 
\label{simresults}
\end{figure*}
A suite of test environments for target search tasks is prepared to benchmark standard algorithms devised over the motion primitive framework.
\subsection{Benchmark Algorithms}

To our knowledge, existing methods do not use motion primitives for agile IPP. Hence, our method is evaluated against some of the traditional IPP algorithms which have been adapted to our motion primitive framework to ensure a fair comparison.
We benchmark our algorithm against five heuristics/algorithms:
\begin{enumerate}
    \item Greedy:
    The greedy algorithm selects the primitive with the highest one-step utility. Here the utility is the informativeness of a certain location (x,y) on the map and is defined as:
    $$
    I(q) = f_1 m_{xy} + f_2 H(m_{xy}) 
    $$
    
    \item Dynamic Programming:
    Dynamic Programming maximizes information gain over a sequence of primitives in a receding horizon manner. Here, the agent executes a single primitive and replans.
    
    \item Covariance Matrix Adaptation Evolutionary Strategy(CMAES):
    CMAES\cite{omidvar2010comparative} samples a sequence of primitives and refines these paths over a fixed horizon to maximize information gain.
    
    \item Online Coverage:
    
    Agents maximize the uniform coverage utility to determine the next motion primitive to execute at a given state. 
    
    \item Online Prioritized Coverage:
    Agents prioritize coverage utility towards regions with higher target beliefs while planning for the next motion primitive. 
    
\end{enumerate}

\begin{table}[t]
\renewcommand{\arraystretch}{1.3}
\caption{Search Performance Comparisons of Baselines over a fixed budget}
\label{table_example}
\centering
\begin{tabular}{l|c|c|c}
{\bf Methods} & \multicolumn{1}{p{1.25cm}}{\centering  \textbf{Coverage (\%)}} & \multicolumn{1}{|p{1.25cm}|}{\centering \textbf{Entropy Reduction (\%)}} &  \multicolumn{1}{p{1.25cm}}{\centering \textbf{Search Efficiency (\%)}}\\
\midrule

Greedy-IPP & 66.73$\pm$19.17&	69.49$\pm$19.77&	68.67$\pm$29.11\\
DP-IPP & 75.77$\pm$14.33&	79.42$\pm$14.08&	79.81$\pm$19.82 \\
CMAES-IPP & 68.95$\pm$5.25 & 73.03$\pm$5.28 & 78.49$\pm$16.87 \\
Coverage & \bfseries79.46 $\pm$11.99 &	78.68 $\pm$13.31	& 78.14$\pm$20.93\\
Prioritised Coverage & 66.76$\pm$9.47	& 68.29$\pm$9.45 & 81.81$\pm$16.98 \\
{\bf Ours-noLSTM} & 74.39$\pm$4.92 &	79.49$\pm$4.28 &	88.91$\pm$9.33 \\
{\bf Ours-LSTM}  & 76.31$\pm$4.04 & \bfseries81.16$\pm$3.83 & \bfseries 90.38$\pm$8.31 \\
\end{tabular}
\end{table}

\subsection{Analysis}

Fig \ref{simresults} shows an example of the traveled path taken by the quad-rotor for target search over a 2D environment.
The left plot confirms that the agent explores the space 
while re-visiting some of the sites to improve estimates that would result in finding targets.

Table 1 shows the performance of the algorithms averaged out over all test maps and we conclude that the performance of our method outperforms the baseline heuristics implemented in terms of search performance in terms of target search efficiency over a large variety of maps.
In each map the agents prior varies from the true ground truth distribution from which targets are initialized.
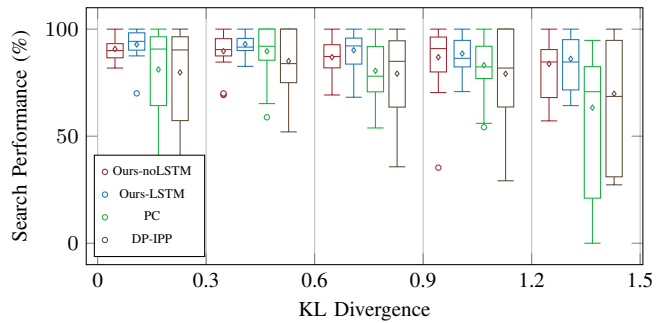
\begin{figure}[htbp]
    \input{./data/plot_div_results_new}
    \caption{%
    Robustness of our approach with varying distributions of the prior: The prior information of the environment can be inaccurate when compared to the true target distribution, as indicated by the KL divergence ($x$-axis).
    Our approaches ({\em red}, {\em blue}) generally perform better than prioritized coverage (PC) method ({\em green}) and the DP-IPP method ({\em brown}) in terms of search efficiency ($y$-axis).
    Furthermore, our approach has smaller deviations from the mean, indicating that the performance of our approach is robust to varying divergence of the prior from the true target distribution.%
    \label{fig:div}}
\end{figure}

In the case of CMAES-IPP, limited online planning results in poorer trajectory quality due to the lack of adaptation to the current state of the target occupancy map. These results also reveal that our method outperforms DP-IPP based methods by sacrificing coverage to improve target estimates in certain regions and outperforms prioritized coverage to explore more areas to find targets. 

This result is also substantiated from Fig \ref{fig:div}, where we contrast the performance of our approach, prioritized coverage and DP-IPP with respect to the KL divergence between the ground truth distribution and the agent's prior.

%% file: data/plot_div_results_new.tex
\definecolor{mBlue}{HTML}{1F77B4}
\definecolor{mDarkBrown}{HTML}{604c38}
\definecolor{mDarkTeal}{HTML}{23373b}
\definecolor{mSteelGray}{HTML}{2B2E34}
\definecolor{mLightBrown}{HTML}{EB811B}
\definecolor{mLightGreen}{HTML}{14B03D}
\definecolor{mDarkRed}{HTML}{a2282f}

\usetikzlibrary{pgfplots.statistics}
    \pgfmathdeclarefunction{fpumod}{2}{%
        \pgfmathfloatdivide{#1}{#2}%
        \pgfmathfloatint{\pgfmathresult}%
        \pgfmathfloatmultiply{\pgfmathresult}{#2}%
        \pgfmathfloatsubtract{#1}{\pgfmathresult}%
        \pgfmathfloatifapproxequalrel{\pgfmathresult}{#2}{\def\pgfmathresult{4.9}}{}%
    }
	\begin{tikzpicture}
	\footnotesize
		\pgfplotsset{
			my boxplot style/.style={
				boxplot,
				solid,
				fill=white,
				mark=*,
				mark size=1.0pt,
				every mark/.append style={
					fill=white,
				},
				},
		}
		\makeatletter
		\pgfplotsset{
			boxplot/draw/average/.code={%
				\draw[/pgfplots/boxplot/every average/.try]
				\pgfextra
				\pgftransformshift{%
					\pgfplotsboxplotpointabbox
					{\pgfplotsboxplotvalue{average}}
					{0.5}%
				}%
				\pgfuseplotmark{\tikz@plot@mark}%
				\endpgfextra
				;
			},
			boxplot/every average/.style={
			/tikz/mark=diamond*,
			},
		}
		\makeatother
		\begin{axis}[
			boxplot,
			boxplot/draw direction=y,
			xtick={0,1,2,3,4,5},
			xticklabels={0,0.3,0.6,0.9,1.2,1.5},
			enlarge y limits,
			enlarge x limits=0.04,
			boxplot/average={auto},
			ylabel = {Search Performance (\%)},
			xlabel = {KL Divergence} ,
			ylabel near ticks,
			grid=major,
			ymajorgrids=false,
			boxplot={
			draw position={0.16 + 1.00*floor(\plotnumofactualtype/4) + 0.20*(fpumod(\plotnumofactualtype, 4))},
			box extend=0.15,
			whisker extend=0.15,
			},
			height=5cm,
			width=1.04\columnwidth,
			label style={font=\footnotesize},
			legend entries = {Ours-noLSTM, Ours-LSTM, PC, DP-IPP},
			legend to name={legend},
			legend style={font=\tiny},
			name=border
			]
			\addplot[my boxplot style,draw=mDarkRed]table[y=Ours-noLSTM, col sep=comma]{./data/results0.csv};
			\addplot+[my boxplot style,draw=mBlue]table[y=Ours-LSTM, col sep=comma]{./data/results0.csv};
			\addplot+[my boxplot style,draw=mLightGreen]table[y=PC, col sep=comma]{./data/results0.csv};
			\addplot+[my boxplot style,draw=mDarkBrown]table[y=DPIPP, col sep=comma]{./data/results0.csv};
			\addplot+[my boxplot style,draw=mDarkRed]table[y=Ours-noLSTM, col sep=comma]{./data/results1.csv};
			\addplot+[my boxplot style,draw=mBlue]table[y=Ours-LSTM, col sep=comma]{./data/results1.csv};
			\addplot+[my boxplot style,draw=mLightGreen]table[y=PC, col sep=comma]{./data/results1.csv};
			\addplot+[my boxplot style,draw=mDarkBrown]table[y=DPIPP, col sep=comma]{./data/results1.csv};
			\addplot+[my boxplot style,draw=mDarkRed]table[y=Ours-noLSTM, col sep=comma]{./data/results2.csv};
			\addplot+[my boxplot style,draw=mBlue]table[y=Ours-LSTM, col sep=comma]{./data/results2.csv};
			\addplot+[my boxplot style,draw=mLightGreen]table[y=PC, col sep=comma]{./data/results2.csv};
			\addplot+[my boxplot style,draw=mDarkBrown]table[y=DPIPP, col sep=comma]{./data/results2.csv};
			\addplot+[my boxplot style,draw=mDarkRed]table[y=Ours-noLSTM, col sep=comma]{./data/results3.csv};
			\addplot+[my boxplot style,draw=mBlue]table[y=Ours-LSTM, col sep=comma]{./data/results3.csv};
			\addplot+[my boxplot style,draw=mLightGreen]table[y=PC, col sep=comma]{./data/results3.csv};
			\addplot+[my boxplot style,draw=mDarkBrown]table[y=DPIPP, col sep=comma]{./data/results3.csv};
			\addplot+[my boxplot style,draw=mDarkRed]table[y=Ours-noLSTM, col sep=comma]{./data/results4.csv};
			\addplot+[my boxplot style,draw=mBlue]table[y=Ours-LSTM, col sep=comma]{./data/results4.csv};
			\addplot+[my boxplot style,draw=mLightGreen]table[y=PC, col sep=comma]{./data/results4.csv};
			\addplot+[my boxplot style,draw=mDarkBrown]table[y=DPIPP, col sep=comma]{./data/results4.csv};
			
		\end{axis}
		\node[above right] at (border.south west) {\ref{legend}};
	\end{tikzpicture}

%% file: conclusion.tex
This paper proposes a novel RL-based adaptive Informative Path Planning with policies defined over dynamic motion primitives of a fast moving robot to achieve target search. The proposed algorithm enables fast information gathering for active classification tasks by planning dynamically feasible agile paths. A key component in our approach is the use of motion primitives to parameterize trajectories which respect UAV's dynamics. The learned policy is validated by comparing its performance to multiple IPP based benchmarks adapted over motion primitives and improves target search efficiency without compromising on runtime constraints, and adaptive replanning to balance between exploration and exploitation. Future work would investigate this planning approach for 3D UAV planning with real world experiments for experimental validation, and UAV teams  .